\documentclass{article}

\usepackage{arxiv}

\usepackage[utf8]{inputenc} % allow utf-8 input
\usepackage[T1]{fontenc}    % use 8-bit T1 fonts
\usepackage{hyperref}       % hyperlinks
\usepackage{url}            % simple URL typesetting
\usepackage{booktabs}       % professional-quality tables
\usepackage{amsfonts}       % blackboard math symbols
\usepackage{nicefrac}       % compact symbols for 1/2, etc.
\usepackage{microtype}      % microtypography
\usepackage{lipsum}		% Can be removed after putting your text content
\usepackage{graphicx}
\usepackage{natbib}
\usepackage{doi}
\usepackage{subfloat}
\usepackage{subcaption}

\title{GENERATIVE PRE-TRAINED TRANSFORMERS FOR BIOLOGICALLY INSPIRED DESIGN}

\date{March 31, 2022}	% Here you can change the date presented in the paper title
\author{ \href{https://orcid.org/0000-0002-5401-6679}{\includegraphics[scale=0.06]{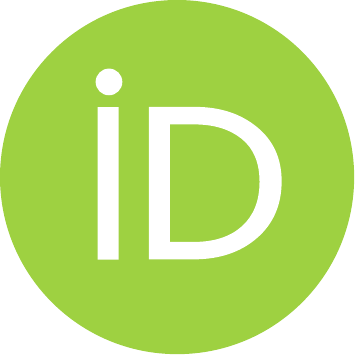}\hspace{1mm}Qihao Zhu}\\
	Data-Driven Innovation Lab\\
	Singapore University of Technology and Design\\
	\texttt{qihao\_zhu@mymail.sutd.edu.sg} \\
	\And
	\href{}{\hspace{1mm}Xinyu Zhnag} \\
	State Key Laboratory of Automotive Safety and Energy\\
	Tsinghua University\\
	\texttt{xyzhang@tsinghua.edu.cn} \\
	\And
	\href{https://orcid.org/0000-0001-5892-8432}{\includegraphics[scale=0.06]{orcid.pdf}\hspace{1mm}Jianxi Luo} \\
	Data-Driven Innovation Lab\\
	Singapore University of Technology and Design\\
	\texttt{luo@sutd.edu.sg} \\
}

\renewcommand{\shorttitle}{GPT for Biologically Inspired Design}

\begin{document}
\maketitle
\begin{abstract}
Biological systems in nature have evolved for millions of years to adapt and survive the environment. Many features they developed can be inspirational and beneficial for solving technical problems in modern industries. This leads to a novel form of design-by-analogy called bio-inspired design (BID). Although BID as a design method has been proven beneficial, the gap between biology and engineering continuously hinders designers from effectively applying the method. Therefore, we explore the recent advance of artificial intelligence (AI) for a computational approach to bridge the gap. This paper proposes a generative design approach based on the pre-trained language model (PLM) to automatically retrieve and map biological analogy and generate BID in the form of natural language. The latest generative pre-trained transformer, namely GPT-3, is used as the base PLM. Three types of design concept generators are identified and fine-tuned from the PLM according to the looseness of the problem space representation. Machine evaluators are also fine-tuned to assess the correlation between the domains within the generated BID concepts. The approach is then tested via a case study in which the fine-tuned models are applied to generate and evaluate light-weighted flying car concepts inspired by nature. The results show our approach can generate BID concepts with good performance.
\end{abstract}

% keywords can be removed

\section{INTRODUCTION}
Bio-inspired design (BID) is a relatively novel design methodology in the engineering design field. It is based on the observation of biological natural phenomena \citep{ISO266} and aims to develop novel solutions for real-world problems with analogies from nature \citep{helms2009biologically}. For millions of years, creatures on earth managed to evolve features to survive the environments and compete with other species, and some of them could be crucial for solving the technical challenges we are facing today. In a bullet train example \citep{linic2020experimental}, engineers observed that kingfishers are capable of diving into the water from air with extremely high speed without making a splash and took the kingfisher’s beak shape as an analogy to develop the head of the bullet train. This improvement not only removed the sonic boom caused by the train traveling at high speed through a tunnel, but also resulted in a faster speed and lower energy consumption.\par
Although the method has been proven helpful and become popular among designers and researchers, it is challenging for designers to find, recognize, and understand biological information in design practice because they lack the necessary knowledge and experience in the biology field \citep{vattam2013information,kruiper2018towards}. For such challenges related to knowledge and information process, a data-driven approach and knowledge-based artificial intelligence can aid human designers for BID \citep{jiang2022data}. Recent advances in deep learning-based show particular promises if we can properly apply them to biological data and design problems.\par
In this paper, we apply one of the latest natural language processing (NLP) techniques to help designers innovate with BID. We propose a novel approach using the pre-trained language model (PLM) to automate the retrieval and mapping of biological information through text generation. The generated concepts can be evaluated through classifiers fine-tuned from the same base PLM. This could generate and evaluate many BID concepts in a very short period, and the natural language representation of the results are understandable for designers. We further validate the approach in a flying car design project and tested its performance to solve a real-world technical problem together with a team of flying car experts.

\section{BACKGROUND AND RELATED WORKS}
\label{sec:headings}
\subsection{Design-by-Analogy}
Design-by-analogy (DbA) is a design methodology that innovates the design target through the analogy with a different domain \citep{goel1997design}, which is proven effective to help overcome design fixation \citep{linsey2010study}. Analogy in design is generally defined as “illustration of an idea by means of another familiar idea that is similar or parallel to it in some significant features” \citep{hey2008analogies}. This indicates two underlying domains in analogy reasoning. So far, researchers have identified these two domains in DbA, the domain that contains the problem and needs to be understood is called the target domain, while the one that provides a potential solution to the problem is called the source domain \citep{goel1997design, hey2008analogies, linsey2008modality}. \par
Many prior studies have investigated the cognitive process and reasoning behind DbA, and they basically agree on three important subphases: retrieval, mapping, and evaluation \citep{hey2008analogies,linsey2008modality,kokinov2003computational, hall1989computational}. The retrieval of source for analogy depends on the relational similarity between domains \citep{linsey2008modality, gentner1997structure, verhaegen2011identifying}. This means that the retrieved source domain should share a common relationship with the target domain regarding how they are comprised by their components. Then, mapping between the target and the retrieved source is established and evaluated. This is usually done by comparing the elements and patterns between domains \citep{holyoak1996thagard,bhatta1996design}.\par 
In recent years, many data-driven design-by-analogy methods and tools have been proposed. Readers may refer to Shuo et al. \citep{jiang2022data} for the latest and comprehensive review of data-driven design-by-analogy research. In this study, we focus on bio-inspired design (BID), which is a special form of DbA, and explore a pre-trained language model to automate the retrieval, mapping, and evaluation of biological analogies for design concept generation.

\subsection{Bio-Inspired Design}
Bio-inspired design (BID) is an innovation method whereby designers take inspiration from biological phenomena. Although some researchers argue that BID goes beyond analogy \citep{ISO266,vincent2006biomimetics,shu2011biologically}. The methodology has grown into an innovative design process in the field of engineering design which leverages analogies to biological systems to develop solutions for engineering challenges \citep{helms2009biologically}. \par
According to literatures, there are two main approaches for bio-inspired design: problem-driven and solution-based \citep{fayemi2017biomimetics, badarnah2015methodology}. The problem-driven approach is a biomimetic development method that aims to address a practical problem, with the problem serving as the process's starting point, while the solution-based approach seeks inspiration from the observation of nature and allows the identification of a design problem, and therefore the knowledge of a biological system of interest serves as the basis for design. For this paper, we focus on the problem-driven approach as it better fits the general design process model proposed by \citet{pahl2007product}.\par
Researchers have identified three domains in problem-driven BID: problem domain, nature domain, and solution domain \citep{badarnah2015methodology}. Generally, the problem domain and solution domain represent the situations of the target domain before and after analogy mapping, while the nature domain is corresponding to the source domain which indicates the biological features used as analogy source. However, it is universally acknowledged that the transitions between the three domains are challenging due to the distinctions between engineering and biology \citep{kruiper2018towards, salgueiredo2016beyond, vincent2002systematic}. Therefore, many contributions in recent years have explored methods and tools trying to bridge the three domains in BID. \par
The transition from problem domain to nature domain includes the abstraction of the engineering problem and the identification of the biological system \citep{vincent2002systematic, wanieck2017biomimetics}. For abstraction, \citet{nagel2010function} employ the functional model of the desired engineering system for representing the essence of functionality, which is then used to explore biological solutions for inspiration. \citet{cheong2011biologically} propose a natural language processing-based process model that begins with the formulation of original functional terms to represent a problem. For biological system identification, AskNature \citep{deldin2014asknature} groups biological strategies and innovations by function according to the Biomimicry Taxonomy, so that it encodes biological information in engineering terms. Another popular attempt on this issue is to search or push biology keywords based on technical problems. \citet{shu2014natural} explore a natural language approach to search for biology keywords for analogy. \citet{chen2021method} also proposed a method to automatically push keywords based on Composite Correlation Intension values. \par
The gap between nature domain and solution domain is believed to be the result of the challenge of designers’ understandability \citep{kruiper2018towards}. And, the representations of biological systems are key to knowledge transfer in BID process \citep{sartori2010methodology}. To bridge this gap, AskNature \citep{deldin2014asknature} provides biomimicry strategies that help designers develop a deeper understanding of how to apply the information within the nature domain. \citet{sartori2010methodology} developed the SAPPhIRE (State-changes, Actions, Parts, Phenomena, Inputs, oRgans and Effects) guideline to formulate four distinct levels of abstraction at which the transfer between domains could take place. \citet{chen2021structure} proposed a method combining dependency parsing and keyword extraction to extract structure-function knowledge from nature domain.\par

\section{RESEARCH METHOD}
\label{sec:headings}
\subsection{Pre-Trained Language Model (PLM)}
To overcome the gaps, we propose a natural language generation approach to bridge the gap between the three domains (i.e., nature, solution, and problem) in BID and create more understandable textual concepts. This is accomplished by the generation and evaluation of design concepts through the application of the pre-trained language model (PLM).\par
PLMs are language models that have been trained with a large dataset of textual information and can be applied to deal with specific language-related tasks \citep{arslan2021comparison}. For example, the base model of BERT was trained with Wiki and books data that contains over 3.3 billion tokens \citep{kenton2019bert}, and the largest PLM nowadays, namely GPT-3, was trained on a 500 billion tokens dataset \citep{brown2020language}. These huge pre-training datasets offer PLMs not only the capability of understanding human language, but also the knowledge and logic that come with it. \par
PLMs like BERT \citep{kenton2019bert} or GPT \citep{brown2020language,radford2019language} have been increasingly popular in the natural language processing (NLP) field for their state-of-the-art performance in many downstream NLP tasks \citep{duan2020study} including text completion and text classification. There are mainly two available mechanisms for a PLM to perform these NLP tasks: fine-tuning and prompt-based learning. Fine-tuning, first proposed by \citet{hinton2006reducing}, is a technique to re-train the pre-trained model with a small amount of task-specific dataset for the task of interest. The parameters of the base model are updated in the process. Prompt-based learning, on the other hand, makes no change to the base model but can leverage what the model has already learned with simple prompts. This mechanism leads to the few-shot learning technique which requires only a few examples of the desired task as prompt \citep{brown2020language}. \par
In this paper, the PLM we use for concept generation and evaluation is the latest version of generative pre-trained transformer (GPT), i.e., GPT-3 \citep{brown2020language}, and we fine-tune the base model for different tasks in our approach. The reasons that we choose to fine-tune GPT-3 over other models or techniques are as follow: 1. The base model of GPT-3 was pre-trained on much larger dataset than any other PLMs, which makes it more knowledgeable about biology and nature; 2. Fine-tuning can better leverage the BID-related knowledge and logic from the prepared dataset as we assume that previous knowledge of BID is not included in the base PLMs; 3. A large dataset is required to fine-tune a BERT or GPT-2 model which is unobtainable for the BID task, while according to OpenAI \citep{openai}, GPT-3 only needs a few hundred of high-quality samples for fine-tuning; 4. Masked language models like BERT are generally weak at natural language generation, because they can only learn contextual representation of words, namely natural language understanding, but not organize and generate language \citep{duan2020study}, which makes them unsuitable for our generative approach.\par

\label{sec:headings}
\subsection{Dataset}
The dataset we use to fine-tune the GPT-3 model is collected from the innovation section of the AskNature website \citep{deldin2014asknature}. The dataset contains 221 excellent BID samples with each sample represented by textual information of a benefits, applications, the challenge, innovation details, and a biomimicry story (Table 1).\par
\begin{table}[h]
	\caption{Data composition of each innovation sample in the fine-tuning dataset}
	\centering
	\begin{tabular}{p{2.8cm}p{2.2cm}p{9.85cm}}
		\toprule
		%\multicolumn{2}{c}{Part}                   \\
		%\cmidrule(r){1-2}
		Component     & Types of Data     & Description\\
		\midrule
		Benefits  &  Keywords   &  The advantages of the innovation.
\\
		\midrule
		Applications  &  Keywords   &  The applications of the innovation.
\\
		\midrule
        The Challenge  &  Paragraph   &  A statement of problem in the target domain that is challenging to solve.
\\
		\midrule
        Innovation Details  &  Paragraph   &  Introduction of the innovation that aims to solve the challenge.
\\
		\midrule
		Biomimicry Story  &  Paragraph   &  A biological natural phenomenon that is applicable as analogy for solving the challenge.
\\
		\bottomrule
	\end{tabular}
	\label{tab:table1}
\end{table}
The variety of information in this dataset offers great opportunities to explore different fine-tuning strategies, and this can be done simply through reformulating and customizing the textual data into input-output pairs. For example, given the input-output pair of applications to innovation details, after fine-tuning the model can generate new innovations based on any applications of user’s interest. If we train it the other way around, i.e., use innovation details as input and applications as output, the model can do a very different task that extract keywords of the applications from any input innovation description. By customizing the AskNature dataset and combining different fine-tuned models, different BID related tasks can be implemented.\par

\label{sec:headings}
\subsection{Concept Generation and Correlation Evaluation}
Based on both DbA and BID literatures, we propose a framework (Figure 1) to address our BID concept generation approach. The generation process starts by retrieving a biological system from nature domain based on a given problem space, and the extracted biology information is then mapped to the solution domain. This is done by fine-tuning the GPT-3 model for text completion task. The evaluation process assesses the correlation between domains, i.e., if the generated solution aiming to solve the given problem, and if it has taken inspiration from the biological information. Separate text classifiers are fine-tuned from the GPT-3 base model to assess each correlation.\par
\begin{figure}[h]
	\centering
	\includegraphics[width = 6cm]{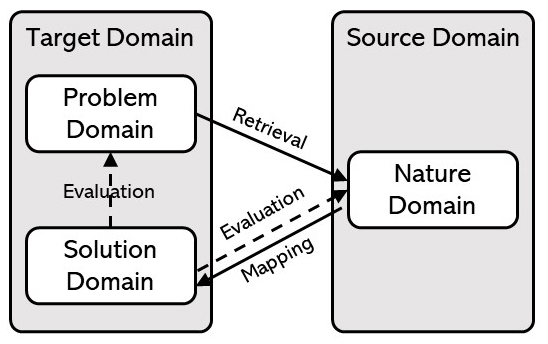}
	%\fbox{\rule[-.5cm]{4cm}{4cm} \rule[-.5cm]{4cm}{0cm}}
	\caption{Theoretical framework}
	\label{fig:fig1}
\end{figure}
For concept generation, we identify three types based on the looseness of the problem space (Table 2). Type-1 generation has the minimum constraints regarding how the problem is defined, it takes only the applications as input and generate biomimicry and innovation based on it. This means that the retrieval and mapping of the biological information only take the target’s applications into consideration and leaves the problem space wide-open. This type is suitable for scenarios where companies want to innovate a certain product of interest, i.e., drone, but without a specific problem in mind. By hypothesis, Type-1 generator should be able to provide the most diverse results.\par
Type-2 constrains the problem space by introducing benefits, i.e., what advantages is this innovation expected to bring. It can be seen as an abstraction of a complex problem into a few keywords of benefits and applications. This type using the abstract representation of the problem is suitable when designers have only a general requirement based on the user’s needs, e.g., a small and light-weighted drone. \par
Type-3 generation further constrains the problem space by using a paragraph of challenge statement, which could contain three to four sentences that describes a problem from the engineering perspective. As the strictest way of problem definition that we can use based on the AskNature dataset, this type could be useful for the engineering teams that may come across a specific and detailed problem in design practice and wish to solve it by BID.\par

\begin{table}[h]
	\caption{Fine-tuned models for generation and evaluation}
	\centering
	\begin{tabular}{p{1.8cm}p{2cm}p{3cm}p{7cm}}
		\toprule
		%\multicolumn{2}{c}{Part}                   \\
		%\cmidrule(r){1-2}
		Model     & Input     & Output & Description\\
		\midrule
		\multicolumn{4}{c}{Generation} \\
		\midrule
		Type-1\par Generation  &  Applications   &  Biomimicry Story, \par Innovation   & Generate concept with an open problem space
\\
		\midrule
		Type-2\par Generation  &  Benefits, \par Applications   &  Biomimicry Story,\par Innovation   &  Represent the problem in simple keywords of applications and benefits
\\
		\midrule
		Type-3\par Generation  &  Challenge Statement   &  Biomimicry Story,\par Innovation   &  Use the full challenge statement as problem space
\\
		\midrule
		\multicolumn{4}{c}{Evaluation} \\
		\midrule
        Problem-Solution Correlation  &  Benefits,\par Innovation   &   Related/Unrelated   &  Evaluate if the generated innovation may bring the given benefits, applicable to type-2 generation
\\
		\midrule
        Problem-Solution Correlation  &  Challenge,\par Innovation   &   Related/Unrelated   &  Evaluate if the generated innovation aims to solve the given challenge, applicable to type-3 generation
\\
		\midrule
        Nature-Solution Correlation  &  Biomimicry,\par Innovation   &   Related/Unrelated   &  Evaluate if the generated innovation takes inspiration from the biomimicry, applicable for all three types of generation
\\
		\bottomrule
	\end{tabular}
	\label{tab:table1}
\end{table}
Due to the varied problem representation of the three types of generation, different sets of classifiers are required to evaluate the correlation between domains. This includes benefits-solution correlation, challenge-solution correlation, and nature-solution correlation (Table 2). However, directly learning relations of logic between texts is difficult for today’s NLP techniques. Therefore, we transform these relation learning tasks into simpler binary text classification tasks to predict if the two domains in the given text information are correlated. To construct the dataset for fine-tuning a classifier, we need negative samples that do not acquire the correlation of interest but maintain the form and topic that are as close to the positive samples as possible. \par
Figure 2 represents our method to construct such datasets. Assuming two domains A and B are to be evaluated if they are correlated. The two domains that form the positive samples are from the collected AskNature dataset, with a label that indicates their relevancy. On the other hand, in negative ones we replace only the B domain into random samples to create the irrelevance. Such a sample is randomly generated by GPT-3. Some techniques are needed to control its form and topic so that the only difference between the positive B domain and the negative B domain is their relevancy to the domain A. \par
The technique to generate negative samples varies based on the domains to be evaluated. For the problem-solution correlation evaluators, an extra generator is fine-tuned from the GPT-3 base model that takes in applications as input and generates random innovation details. This ensures the negative solution samples ignore the problem, but still describe the same product application in the same form as in AskNature. Similarly, the negative samples for nature-solution correlation are also prepared by random generation, whereas this time we select five samples of innovation details on AskNature that do not contain any biological information and use them as examples for few-shot learning in GPT-3. Negative samples prepared in this technique contain no biological information. This is because we observed that the bad generation results usually didn’t take the wrong nature domain as inspiration, but they do not use analogy from nature at all. \par
Furthermore, as \citet{soares2019matching} suggest for language model to learn relations, marker tokens are added with all samples for fine-tuning the classifiers. Marker tokens are simple tags that are added before and after the sentences or paragraphs of interest. An example of how the marker tokens are used is: \textit{“[Bio]Octopus tentacles have suckers that allow the organisms to hold small objects. The suckers have small…[/Bio][Inno]The soft manipulator was inspired by the suckers of octopus tentacles. It is made of a temperature-responsive layer of…[/Inno]”}. The tags that come with square brackets in the above example are marker tokens.\par
\begin{figure}[h]
	\centering
	\includegraphics[width = 15cm]{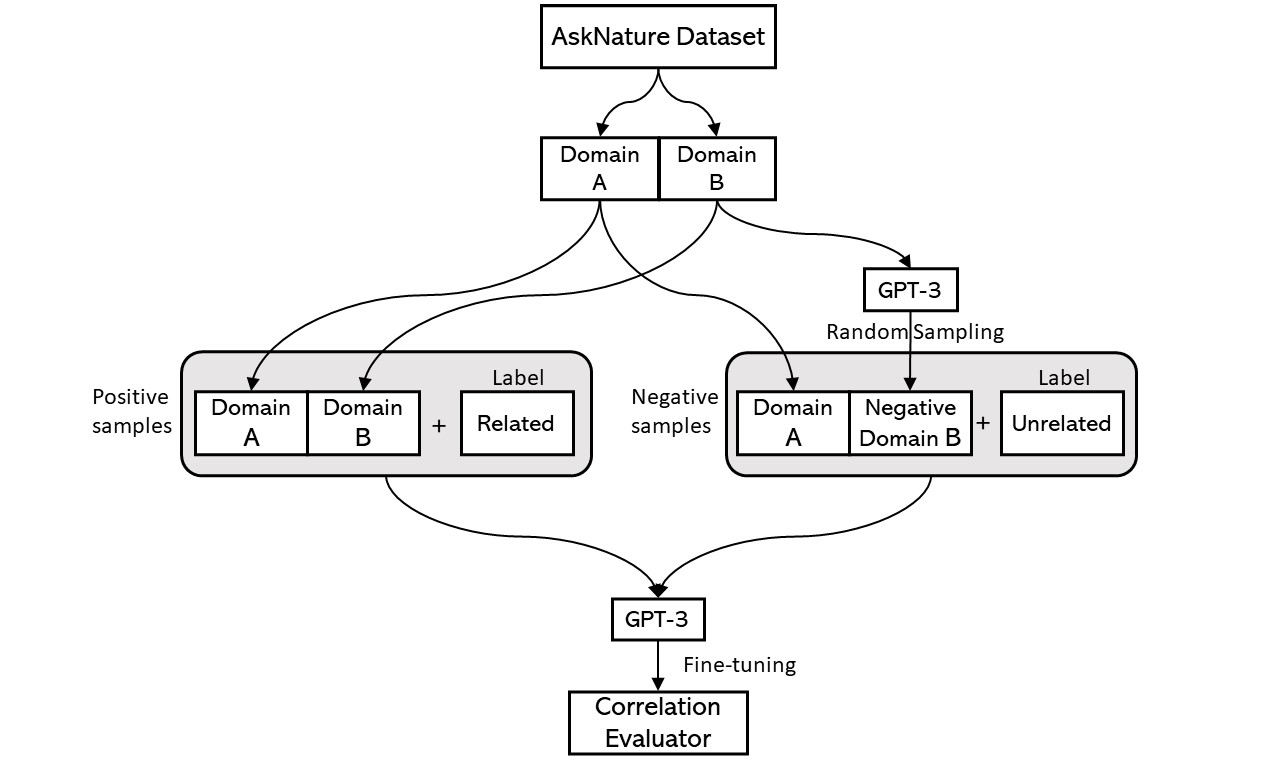}
	%\fbox{\rule[-.5cm]{4cm}{4cm} \rule[-.5cm]{4cm}{0cm}}
	\caption{Dataset construction method for fine-tuning the correlation evaluators}
	\label{fig:fig2}
\end{figure}

\section{RESULTS AND DISCUSSION}
\label{sec:headings}
\subsection{Fine-tuning Results}
The three types of generators are fine-tuned from GPT-3’s base model of Davinci, which is currently the largest base model that OpenAI provides via their API. The model was fine-tuned for 4 epochs with the batch size of 1 (batch size is determined based on 0.2\% of the size of training samples). The gradually decreasing training loss of all three types illustrated in figure 3 shows the model’s capability of learning BID.\par
\begin{figure}[h]
	\centering
	\includegraphics[width = 12cm]{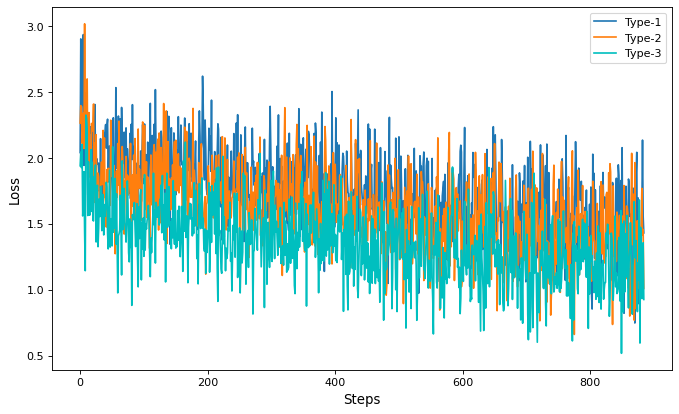}
	%\fbox{\rule[-.5cm]{4cm}{4cm} \rule[-.5cm]{4cm}{0cm}}
	\caption{Training loss for fine-tuning three types of concept generation}
	\label{fig:fig3}
\end{figure}
To assess the models regarding their capability of generating novel design concepts, we employ the metrics of Word Mover’s Distance (WMD) \citep{kusner2015word}. WMD is widely applied in NLP field to measure the distance between two text documents based on word embeddings and is popular in natural language generation (NLG) tasks like image captioning \citep{kilickaya2017re} and machine translation \citep{chow2019wmdo} to assess the performance of the NLG model. Such applications often aim for minimizing WMD as they want the generated results to be as close to the ground truth as possible. However, we want to do the opposite in our design concept generation task, i.e., assess which model gives the larger WMD, because it means the model can generate more diverse results. In design tasks, the diversity of concepts could lead to novel solutions, and therefore we use the metrics to represent the model’s capability of generating potentially novel concepts. Note that this approach can only assess the performance of the generator model but not the novelty of an individual concept generated with customized input because the ground truth for similarity measurement is needed and it only exists in samples in the dataset.\par
To measure the WMD, we select three samples of robotics from the AskNature innovation dataset: an aerial robot, a ground robot, and an underwater robot. Their applications, benefits, and challenges are used as input according to the three types of generation, and 50 concepts are generated for each sample based on each generator type. Then, we extract the word embeddings of the innovation details of each original sample and generated ones, measure the embeddings similarity using WMD. The word embedding extraction method we use in this experiment is the pre-trained Word2Vec model provided by Gensim (https://radimrehurek.com/gensim/). Figure 4 (a) (b) (c) shows the WMD distribution of the 50 results of each sample and each type, figure 4 (d) organizes all the 450 results based on types. Type-1 generation results in the most diverse concepts but only slightly better than Type-2, while Type-3 gives the least diverse ones. This is not surprising as Type-1 and Type-3 have the loosest and strictest defined problem space as constraints, respectively.\par
\begin{figure}[h]
	\centering
	\includegraphics[width = 12cm]{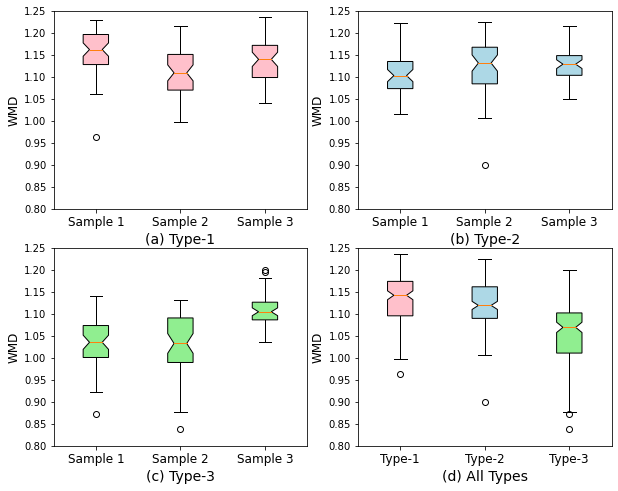}
	%\fbox{\rule[-.5cm]{4cm}{4cm} \rule[-.5cm]{4cm}{0cm}}
	\caption{WMD distribution of generated concepts}
	\label{fig:fig4}
\end{figure}
The evaluators are fine-tuned from GPT-3’s base model of Curie, which is not as strong as Davinci but quite capable of complex text classification [38]. Figure 5 shows the classification accuracy when fine-tuning the evaluators. The accuracy of the benefits-innovation evaluator and challenge-innovation evaluator reach 67.4\% and 79.8\% respectively after 4 epochs, while the biomimicry-innovation evaluator achieves 97.8\% accuracy after only 1 epoch.\par
However, the 67.4\% accuracy of benefits-innovation correlation evaluator is acceptable but not very good, which may have negative influences on the evaluation of Type-2 generation results. The authors will continue working on improving the classification accuracy in future works.\par
\vskip 0.2cm

\begin{figure}[h]
	\centering
	\includegraphics[width = 8cm]{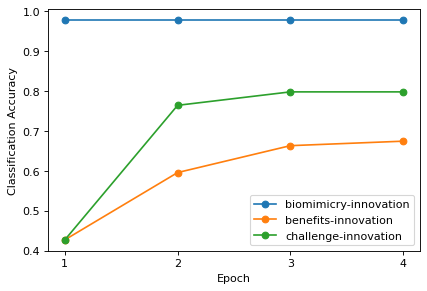}
	%\fbox{\rule[-.5cm]{4cm}{4cm} \rule[-.5cm]{4cm}{0cm}}
	\caption{Classification accuracy of the three correlation evaluators}
	\label{fig:fig4}
\end{figure}

\label{sec:headings}
\subsection{Case Study}
We have applied our BID concept generation approach in a real government-funded project to design flying cars. Along with the acceleration of the urbanization process, ground traffic in cities is crowded and shows increasing pressure. However, the airspace above the urban area has not yet been developed. Air transportation becomes a prospective solution for road congestion. To merge the air and ground traffic, the flying car is proposed as a class of novel dual-mode vehicles. Figure 6 below illustrates three well-known flying car prototypes. However, many challenges still remain to be solved before we can finally see them flying and driving in the cities. \par
\begin{figure}[h]
	\centering
	\includegraphics[width = 10cm]{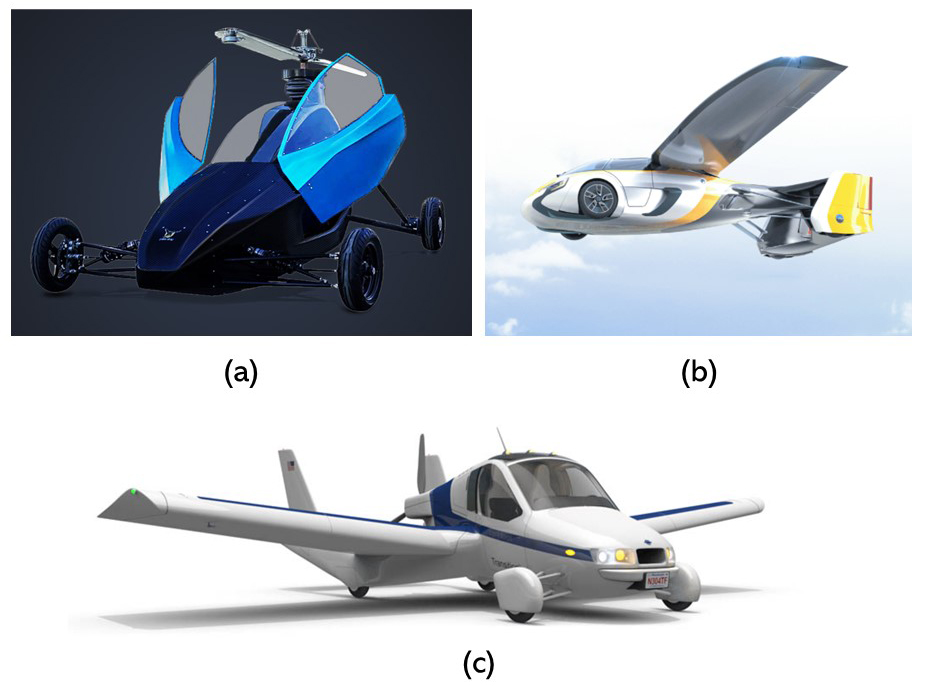}
	%\fbox{\rule[-.5cm]{4cm}{4cm} \rule[-.5cm]{4cm}{0cm}}
	\caption{Flying car examples: (a) Pegasus E-Model (https://bepegasus.com/index.jsp) (b) Aeromobil (https://www.aeromobil.com/) (c) Terrafugia Transition (https://terrafugia.com/transition/)
}
	\label{fig:fig4}
\end{figure}
At the beginning of the project, we interviewed the project team of 10 engineers that had worked on several flying car design projects regarding the challenges they came across in their design projects including the present one. Lightweight design emerges as a major design problem of flying cars and requires novel solutions. Table 3 lists the corresponding inputs to all three types of fine-tuned GPT-3 generators (Table 2). For each type, 50 concepts are generated.\par
\begin{table}[h]
	\caption{Inputs of flying car case study}
	\centering
	\begin{tabular}{p{2cm}p{9cm}}
		\toprule
		%\multicolumn{2}{c}{Part}                   \\
		%\cmidrule(r){1-2}
		Applications     & Flying car\\
		\midrule
		Benefits  &  Lightweight\\
		\midrule
		Challenge Statement  &  A flying car includes a subsystem for flying in the air in addition to a subsystem for driving on the ground. With both flying and driving subsystems in one, the weight might be increased to increase the drive load, demand more propulsion to overcome gravity, and increase fuel consumption. Lightweight design is a challenge for flying cars.\\
		\bottomrule
	\end{tabular}
	\label{tab:table3}
\end{table}
Figure 7 categorizes the biological sources of the GPT-generated bio-inspired flying car design concepts. Across all three types, the generators are over 50\% likely to retrieved birds as an analogy, and then insects. This is consistent with the cognitive process of humans where people tend to retrieve information based on attribute similarities \citep{linsey2008modality}, i.e., retrieve flying animals as analogies for flying cars. However, there are some unexpected but interesting creatures like reptiles and even plants, which could result in more novel concepts.\par
\begin{figure}[h]
	\centering
	\includegraphics[width = 15cm]{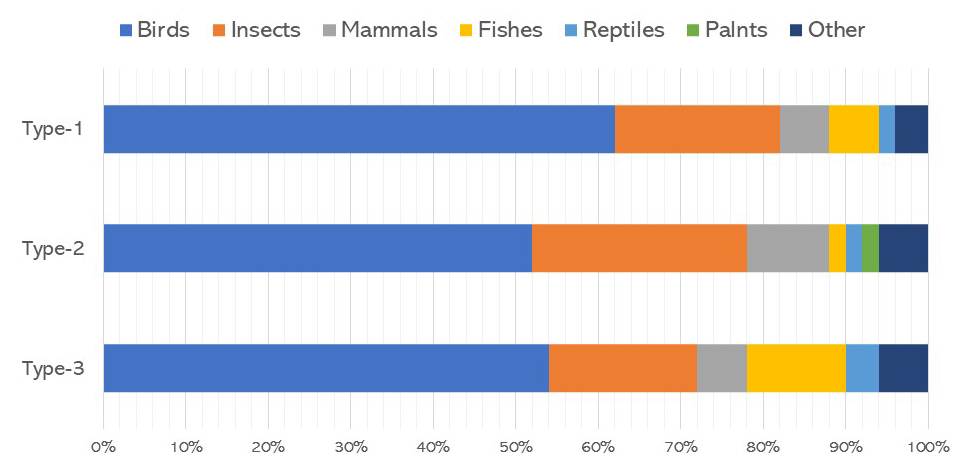}
	%\fbox{\rule[-.5cm]{4cm}{4cm} \rule[-.5cm]{4cm}{0cm}}
	\caption{Categories of the retrieved biological systems}
	\label{fig:fig7}
\end{figure}
The correlation evaluation results are shown in Table 4. Type-1 is excluded for the evaluation of problem-solution correlation because the problem space is open. We evaluated the remaining GPT-generated BID concepts passing the tests and narrow down to a small number of candidates for embodiment design, detailed design, and physical prototyping. We chose two concepts generated by each type and asked the flying car design team to evaluate them in two dimensions of feasibility and novelty.\par
\begin{table}[h]
	\caption{Percentage of generated concepts passing the evaluations}
	\centering
	\begin{tabular}{p{3cm}p{2cm}p{2cm}p{2cm}}
		\toprule
		%\multicolumn{2}{c}{Part}                   \\
		%\cmidrule(r){1-2}
		& Type-1 & Type-2 & Type-3 \\
		\midrule
		Problem-Solution     & N/A & 44\% & 84\% \\
		\midrule
		Nature-Solution     & 86\% & 90\% & 86\% \\
		\midrule
		Overall             & 86\% & 40\% & 76\% \\
		\bottomrule
	\end{tabular}
	\label{tab:table4}
\end{table}
Table 5 lists the concepts for the team to evaluate. Two concepts from the Type-1 generator that passes the evaluation are also included. These can provide an intuitive sense to the readers on the GPT-generated concepts as well as the performance of our approach. The model is capable of learning how to retrieve biological information that shares attributional or relational similarity with the target, and then map it to the target to generate solutions. For instance, the second concept in Table 5 proposes to mimic the lightweight skeleton of pterosaurs to design the flying car body. Meanwhile, both biomimicry and innovation descriptions could be not concise enough and contain extra information that is irrelevant to the analogy of our interest.\par
Moreover, we also added two samples of robotics innovations from AskNature as the benchmarking group for the tam to evlaute. They are: “Safer aerial robot inspired by pufferfish” (https://asknature.org/innovation/safer-aerial-robot-inspired-by-pufferfish/) and “Highly maneuverable drone inspired by swifts” (https://asknature.org/innovation/highly-maneuverable-drone-inspired-by-swifts/). This could make a better sense of how good our generated concepts are compared to the existing BID examples.\par 
\begin{table}[ht!]
\small
	\caption{Concepts for team evaluation}
	\centering
	\begin{tabular}{p{0.2cm}p{15cm}}
		\toprule
		%\multicolumn{2}{c}{Part}                   \\
		%\cmidrule(r){1-2}
		\multicolumn{2}{c}{Type-1 Generation} \\
		\midrule
		1\#     & Biomimicry: Swimming animals must produce thrust to move forward. The thrust is generated by moving the tail fin, which works like a propeller.\par
        Innovation: The flying car was inspired by the way a shark swims through water. The shark tail fin moves back and forth to create thrust, just like a propeller. However, the shark fin has shape-shifting skin inside that helps to smooth out the flow of water while swimming. The flying car has a similar shape-shifting skin inside its tail fin to help with thrust, similarly to the shark fin. It has a rectangular profile when at rest. When it moves through the air, the tail fin smoothly changes shape to a wedge-shape profile, similar to a shark fin. The two different shapes affect the way air moves over the tail fin, creating thrust.
 \\
		\midrule
		2\#     & Biomimicry: Pterosaurs are considered the dinosaurs of the Mesozoic era, because they soared as high as 50 m above the ground. They were the largest flying animals ever, with wingspans of approximately 3 m. Their wings were thin and long, and they had a bony breastbone that was shaped like an upside-down ‘V’. Their forelimbs were short, but their claws were long and sharp, helping them to grab on to branches.\par
        Innovation: The flying car has a body that is similar in shape to pterodactyls, with a body designed to control drag, lift, and thrust. It also has a lightweight hull and a propeller to generate thrust. The vehicle’s hull is constructed of high-performance carbon fiber, inspired by the lightweight skeletons of pterosaurs. The propeller is mounted on a pivoting arm that is controlled by a joystick. The entire assembly weighs approximately 35 pounds and looks similar to a parasail. The propeller is 16 inches in diameter and is powered by a 930cc marine engine.
 \\
        \midrule
        \multicolumn{2}{c}{Type-2 Generation} \\
		\midrule
		3\#             & Biomimicry: Hummingbirds are lightweight, small animals that need to stay aloft while carrying food to their nestlings. They do this by using the forces of lift and drag, which are affected by airspeed and wing shape. At low airspeeds, hummingbird wings provide thrust by moving back and forth. At high airspeeds, their curved outer wing surfaces trap air, creating low drag and enabling them to stay aloft at high speeds.\par
        Innovation: The FLY Bot is made of carbon fiber and epoxy and weighs only 9.4 ounces (266 grams). It is powered by a brushless DC motor and has three wings that are made of a porous, lightweight material. The wings have a curved profile similar to a hummingbird, which allows it to have high lift at low speeds while also being able to move the robot quickly through the air. The high-aspect-ratio wings have porous materials in the inner cores, including carbon fiber, materials that are similar to the sponges found in sand dollars and the nano structure of sea sponges, which allows them to be lightweight as well as strong and flexible. The wings are also connected to close-fitting anchors that are used to control aerodynamic forces. This allows the robot to fly at high speeds of up to 27 miles per hour. Overall, the FLY-BOT weighs only a pound and its wingspan is 10 feet long.
 \\
        \midrule
		4\#             & Biomimicry: Many birds are able to fly for very long periods of time without tiring. This is because birds are able to keep their heart rates low and oxygen consumption at a minimum. Their lightweight and porous bones allow their bodies to keep warm, which helps to conserve energy.\par
        Innovation: The lightweight carbon fiber body of the flying car has many of the same porous and lightweight bone characteristics found in a bird. It has a carbon fiber body with a porous, lightweight lattice of carbon fiber ribs that support the vehicle and minimize drag while in flight. The ribs are attached via carbon fiber spars that are held together by a unique adhesive that was inspired by gecko feet, which allow the vehicle to stay attached when flying through the air.
\\
        \midrule
        \multicolumn{2}{c}{Type-3 Generation} \\
        \midrule
		5\#             & Biomimicry: Some insects, such as dragonflies and fruit flies, are known for their high maneuverability and high speed flight. This agility is enabled by a wing geometry that gives the wing a high aspect ratio (length to width ratio) and a thin structure.\par
        Innovation: The wing design of the flying car is inspired by the wing shape of dragonflies and fruit flies. It is made of lightweight carbon fiber that has the same stiffness as steel, but is 60\% lighter. The wing is able to effectively integrate air and ground characteristics into a single wing design, which could lead to more efficient use of materials, increased maneuverability, and more. The main wing is wedge-shaped and has a very small aspect ratio, similar to fruit flies and dragonflies. The small aspect ratio significantly reduces the total wing surface area, generating greater lift for a given wing area. The high lift generates significantly less drag and less induced drag, which allows the vehicle to achieve greater efficiency and a better aerodynamic performance.
\\
        \midrule
		6\#             & Biomimicry: The flying fish stores energy for its leap above the water surface. It does this by bending its body in a “U” shape, with its tail lying flat. When it is ready to jump, it springs its tail upwards and propels itself through the air.\par
        Innovation: The lightweight flying car has a horizontal propeller fitted with a folding propeller system, similar to a flying fish. The propeller folds when the vehicle is in driving mode, and unfolds when going into flying mode. When folded, the propeller is protected and the vehicle can travel over land and water.
\\
		\bottomrule
	\end{tabular}
	\label{tab:table4}
\end{table}
The rubrics for feasibility and novelty evaluation are shown in Table 6. For feasibility, we generally refer to the idea of workability by \citet{douglas2006identifying} which represents the ability to implement the concept. On the other hand, novelty is the degree that measures how common the concept has already been seen in the target domain and its related industries. Note that by our metrics, a concept that had been expressed before but was never successfully implemented is still considered novel.
\begin{table}[h]
	\caption{Feasibility and novelty rating rubrics}
	\centering
	\begin{tabular}{p{1.5cm}p{13cm}}
		\toprule
		%\multicolumn{2}{c}{Part}                   \\
		%\cmidrule(r){1-2}
		Feasibility Rank & Description \\
		\midrule
		1     & The concept makes no sense from the engineering perspective. \\
		\midrule
		2     & The concept makes little sense with today’s technology, but could be possible in the future. \\
		\midrule
		3     & The concept makes sense, but efforts are needed to work out a practical technical roadmap. \\
		\midrule
		4     & The concept makes good sense, and a technical roadmap can be easily established to realize it. \\
		\midrule
		5     & The concept makes perfect sense and there are existing tools, materials, or components to realize it. \\
		\midrule
		Novelty Rank & Description \\
		\midrule
		1     & Solution exists and is commonly seen in the target domain (target domain = the target product described in the concept) \\
		\midrule
		2     & Solution exists but is uncommon in target domain. \\
		\midrule
		3     & New features are proposed for target domain, but similar approaches or technology can be commonly seen in related industries (related industries = flying car, drone, automobile, robotics, etc.) \\
		\midrule
		4     & New features are proposed for target domain, and similar approaches or technology can be rarely found in related industries. \\
		\midrule
		5     & New features are proposed, and no similar approaches or technology can be found nowadays. \\
		\bottomrule
	\end{tabular}
	\label{tab:table6}
\end{table}
After the survey, 9 out of the 10 returned results are useable and are analyzed in Figure 8 by the generator types. The team gave relatively high novelty levels but only intermediate feasibility levels to the comparison group, which is reasonable because a lot of samples from AskNature are not directly workable with existing materials or components. Comparing to this, Type-1 and Type-2 concepts got lower feasibility score but higher novelty score, while the ones generated from Type-3 are considered more feasible but less novel. This finding is not surprising because Type-1 and Type-2 generations are more open than Type-3, and this is consistent to our finding from Figure 3 in Section 4.1. \par
Overall, the results show that feasible and novel BID flying car concepts are generated by fine-tuned GPT for the engineering team for selecting the one to move forward for embodiment and detailed design.\par
\begin{figure}[h]
	\centering
	\includegraphics[width = 12cm]{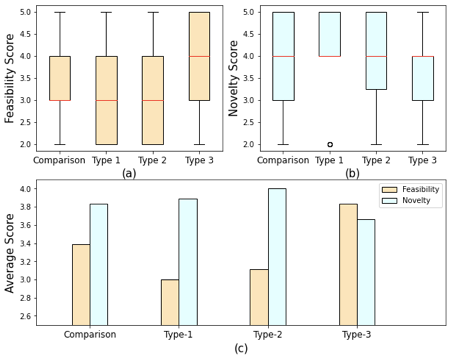}
	%\fbox{\rule[-.5cm]{4cm}{4cm} \rule[-.5cm]{4cm}{0cm}}
	\caption{Human evaluation results: distribution of (a) feasibility scores; (b) novelty scores; (c) average score of feasibility and novelty}
	\label{fig:fig8}
\end{figure}
With the evaluation, we want to select a concept for further development. Concept 4\# received the highest average novelty score among all concepts with an acceptable feasibility score. It suggests mimicking the porous structure of the bone of birds for the material design in the flying car and beyond the expertise of the team. Eventually concept 2\# is selected to proceed for embodiment and detailed design. 2\# draws analogy from the pterosaurs that were the largest known flying animal on earth and yet they can fly fast. The concept suggests mimicking the body and lightweight skeleton of pterosaurs. Figure 9 (a) is the sketch drawn by the first author that illustrates the early embodiment design based on the generated concept. Apart from the body shape and wing structure mentioned in the GPT-generated concept, we also take inspiration from the folding mechanism of pterosaurs’ wings, which results in a combination of the front wheels and the front propellers to further reduce the weight. Figure 9 (b) depicts the folding process from the aerial mode to land mode of the flying car compared to that of the pterosaurs.\par
This interesting example showcases that human-AI collaboration in design concept generation does not necessarily stop at applying what the algorithm may offer. Concepts that are represented in natural language can be either too abstract to understand or too rigid and thus need the human designers to further take them into embodiment and detailed design.  Designers can also draw further inspiration from the given concept and combine inspiration from other sources to make it more suitable, add more details and optimize it for their own project scenario. \par
\begin{figure}
    \centering
    \begin{subfigure}[b]{10cm}
        \includegraphics[width=10cm]{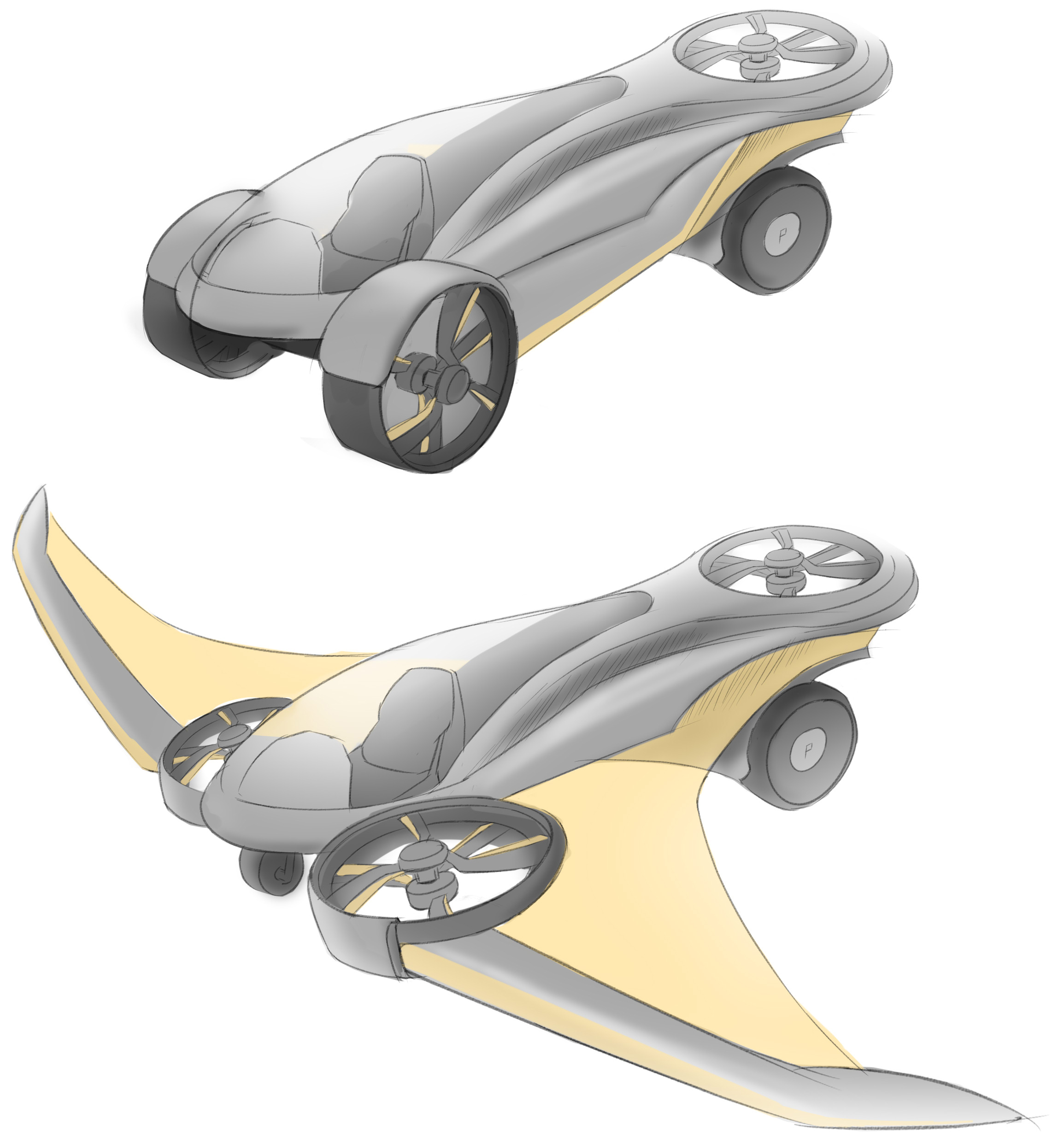}
        \caption{}
        \label{fig:gull}
    \end{subfigure}
    ~ %add desired spacing between images, e. g. ~, \quad, \qquad, \hfill etc. 
    
    \begin{subfigure}[b]{15cm}
        \includegraphics[width=15cm]{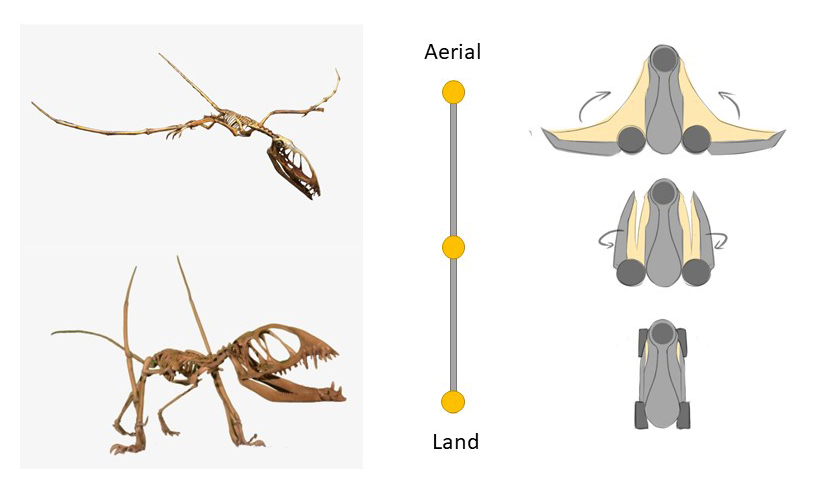}
        \caption{}
        \label{fig:mouse}
    \end{subfigure}
    \caption{Design sketch based on concept 2\#: (a) folding mechanism (b) air-land mode transition}\label{fig:fig9}
\end{figure}

\section{CONCLUDING REMARKS}
This paper has introduced a data-driven intelligent methodology for bio-inspired design concept generation. The core of the methodology is to fine-tune the generative pre-trained transformers for different concept generation situations with different data inputs and use them to automatically generate design concepts in natural language texts. The case study shows reasonable performance of the methodology in generating good quality BID concepts in intelligible natural language. Our work responds to the recent calls in the design research community to adopt pre-trained language models for design research and applications \citep{han2022semantic,siddharth2021natural} To the best of our knowledge, this study is the first to employ generative pre-trained transformers for bio-inspired design, while our recent study has explored the applications of GPT-2 and GPT-3 to generate new design concepts using a product design data repository \citep{zhu2021generative}.\par
Here we aim to contribute to artificial intelligence for BID. Despite being the first to explore fine-tuning GPT-3 to aid in BID concept generation, our work is preliminary and limited in a few aspects. First, we only explored the small AskNature data repository. Future research may explore alternative and even more fruitful nature or nature-inspired design data sources for applying our GPT-based BID methodology. Second, we only introduced three model fine-tuning and concept generation strategies in this paper. There could be more and more flexible strategies for fine-tuning GPT-3 for concept generation. Third, we only explored a limited set of metrics for automatically evaluating the generated concepts in text. Future research may explore more metrics and develop new metrics for evaluating the quality and novelty of the GPT-generated concepts in natural language. In addition, the case study is limited in its scale and scope. The evaluation of 10 engineers from one design project is insufficient to provide statistical significance. As the next step of this research, we plan to engage more and more diverse engineers to evaluate the concepts generated by the fine-tuned GPT for different design problems and settings. These will allow us to develop more systematic understandings on the enabling or conditioning factors on the performance of our methodology.\par
Therefore, these limitations of our work present new research opportunities and future research directions. In general, we hope our exploratory work presented here can inspire more research on the use or development of generative PLM for biologically inspired design, and design-by-analogy in general.\par

\bibliographystyle{unsrtnat}
\bibliography{bibliography-bibtex.bib}  %%% Uncomment this line and comment out the ``thebibliography'' section below to use the external .bib file (using bibtex) .





% \end{thebibliography}

\end{document}